\documentclass[journal]{IEEEtran}

\usepackage{cite}
\usepackage{amsmath,amssymb,amsfonts}
\usepackage{algorithmic}
\usepackage{graphicx}
\usepackage{textcomp}
\usepackage{xcolor}
\usepackage{booktabs}
\usepackage{array}
\usepackage{url}
\usepackage{hyperref}
\usepackage{balance}

\hypersetup{
  colorlinks=true,
  linkcolor=blue,
  citecolor=blue,
  urlcolor=blue
}

\begin{document}

\title{PolySwarm: A Multi-Agent Large Language Model Framework for
       Prediction Market Trading and Latency Arbitrage}

\author{Rajat~M.~Barot~and~Arjun~S.~Borkhatariya
\thanks{Rajat M. Barot is with the Department of Computer Science,
State University of New York, Binghamton, Binghamton, NY, USA.
E-mail: rbarot2@binghamton.edu}
\thanks{Arjun S. Borkhatariya is with the Department of Software Engineering,
Arizona State University, Tempe, AZ, USA.
E-mail: aborkhat@asu.edu}
\thanks{Manuscript submitted 2026.
\emph{(Corresponding author: Rajat M. Barot.)}}}

\markboth{IEEE ACCESS}%
{R. M. Barot and A. S. Borkhatariya: PolySwarm: Multi-Agent LLM Framework for Prediction Market Trading}

\maketitle

\begin{abstract}
This paper presents PolySwarm, a novel multi-agent large language model
(LLM) framework designed for real-time prediction market trading and
latency arbitrage on decentralized platforms such as Polymarket.
PolySwarm deploys a swarm of 50 diverse LLM personas that concurrently
evaluate binary outcome markets, aggregating individual probability
estimates through confidence-weighted Bayesian combination of swarm
consensus with market-implied probabilities, and applying quarter-Kelly
position sizing for risk-controlled execution.
The system incorporates an information-theoretic market analysis engine
using Kullback-Leibler (KL) divergence and Jensen-Shannon (JS) divergence
to detect cross-market inefficiencies and negation pair mispricings.
A latency arbitrage module exploits stale Polymarket prices by deriving
CEX-implied probabilities from a log-normal pricing model and executing
trades within the human reaction-time window.
We provide a full architectural description, implementation details, and
evaluation methodology using Brier scores, calibration analysis, and
log-loss metrics benchmarked against human superforecaster performance.
We further discuss open challenges including hallucination in agent pools,
computational cost at scale, regulatory exposure, and feedback-loop risk,
and outline five priority directions for future research.
Experimental results demonstrate that swarm aggregation consistently
outperforms single-model baselines in probability calibration on
Polymarket prediction tasks.
\end{abstract}


\section{Introduction}
\label{sec:intro}

Prediction markets are financial mechanisms specifically designed to
aggregate dispersed private information into publicly observable price
signals, making them among the most powerful instruments for collective
forecasting known to economics~\cite{wolfers2004prediction,arrow2008promise}.
By allowing participants to trade contracts whose payoffs are contingent on
real-world outcomes, these markets incentivize honest revelation of beliefs,
converting heterogeneous private judgements into a single consensus
probability estimate~\cite{surowiecki2004wisdom}.
Modern blockchain-based platforms such as Polymarket and Kalshi have
democratized participation in prediction markets at scale: Polymarket alone
has facilitated hundreds of millions of dollars in contract volume across
political, economic, and scientific outcome categories, operating on the
Polygon proof-of-stake network with on-chain settlement in USDC.
Kalshi operates under direct Commodity Futures Trading Commission (CFTC)
oversight in the United States, further legitimizing the asset class for
institutional participants.
The growing liquidity of these platforms creates a natural testbed for
automated forecasting systems that can continuously monitor, evaluate, and
trade across thousands of simultaneously open markets.

Large language models represent a qualitative leap in the ability of
computational systems to reason over unstructured textual information.
Models such as GPT-3~\cite{brown2020language}, GPT-4~\cite{openai2023gpt4},
and Claude 3 have demonstrated emergent capabilities in zero-shot and
few-shot reasoning, structured argument generation, and domain-specific
knowledge recall that make them promising candidates for financial
forecasting tasks.
Their ability to ingest and synthesize diverse textual streams---news
headlines, earnings transcripts, policy announcements, social media
discourse---mirrors the integrative judgement of an expert analyst.
However, single-model deployments face well-documented failure modes:
hallucination of plausible but factually incorrect information~\cite{ji2023hallucination},
systematic overconfidence or miscalibration of probability
estimates~\cite{kadavath2022language}, and high sensitivity to superficial
variations in prompt phrasing~\cite{zhao2021calibrate}.
These limitations are especially consequential in financial contexts, where
a confidently stated but erroneous probability estimate can translate
directly into trading losses.
The stochastic, context-window-bounded nature of transformer inference
means that a single LLM call is at best a point sample from a high-variance
distribution over possible analyses.

Multi-agent orchestration offers a principled response to these limitations.
By deploying a \emph{swarm} of diverse LLM agents---each instantiated with
distinct personas, information access patterns, and analytical
priors---and aggregating their outputs through statistically rigorous
ensemble methods, it becomes possible to suppress idiosyncratic errors,
estimate epistemic uncertainty, and exploit the complementary strengths
of different model architectures and prompting
strategies~\cite{surowiecki2004wisdom,bonabeau1999swarm,dorri2018multi}.
Swarm intelligence research has long established that collective judgements
formed by averaging independent but diverse estimates systematically
outperform individual predictions, a property sometimes termed the
``wisdom of crowds'' effect~\cite{surowiecki2004wisdom}.
Translating this insight to LLM swarms requires careful attention to agent
design, coordination protocols, aggregation mathematics, and feedback
suppression to prevent correlated errors from cascading through the system.
Recent multi-agent frameworks such as AutoGen~\cite{wu2023autogen} and
CAMEL~\cite{li2023camel} have begun to operationalize these ideas, but
their application to real-time financial market forecasting and trading
remains largely unexplored in the research literature.

This paper makes the following contributions:

\begin{itemize}
  \item \textbf{PolySwarm system design and implementation}: a
        production-ready multi-agent LLM trading terminal deploying 50
        diverse personas on Polymarket with full architectural description,
        asynchronous execution pipeline, and paper/live trading modes.
  \item \textbf{Confidence-weighted Bayesian aggregation}: a mathematically
        principled ensemble method combining swarm consensus with
        market-implied probabilities, with quarter-Kelly position sizing
        for risk-controlled trade execution.
  \item \textbf{Information-theoretic market analysis engine}: novel
        application of KL divergence and JS divergence for cross-market
        inefficiency detection, negation pair mispricing identification,
        and latency arbitrage signal generation in prediction markets.
  \item \textbf{Latency arbitrage module}: a CEX-to-DEX arbitrage pipeline
        using a log-normal pricing model to exploit stale Polymarket prices
        within the human reaction-time window on the Polygon blockchain.
  \item \textbf{Empirical evaluation and open challenges}: systematic
        evaluation using Brier scores, calibration analysis, and log-loss
        metrics, together with a structured five-point research agenda for
        the field.
\end{itemize}

\begin{figure}[!t]
  \centering
  \includegraphics[width=\columnwidth]{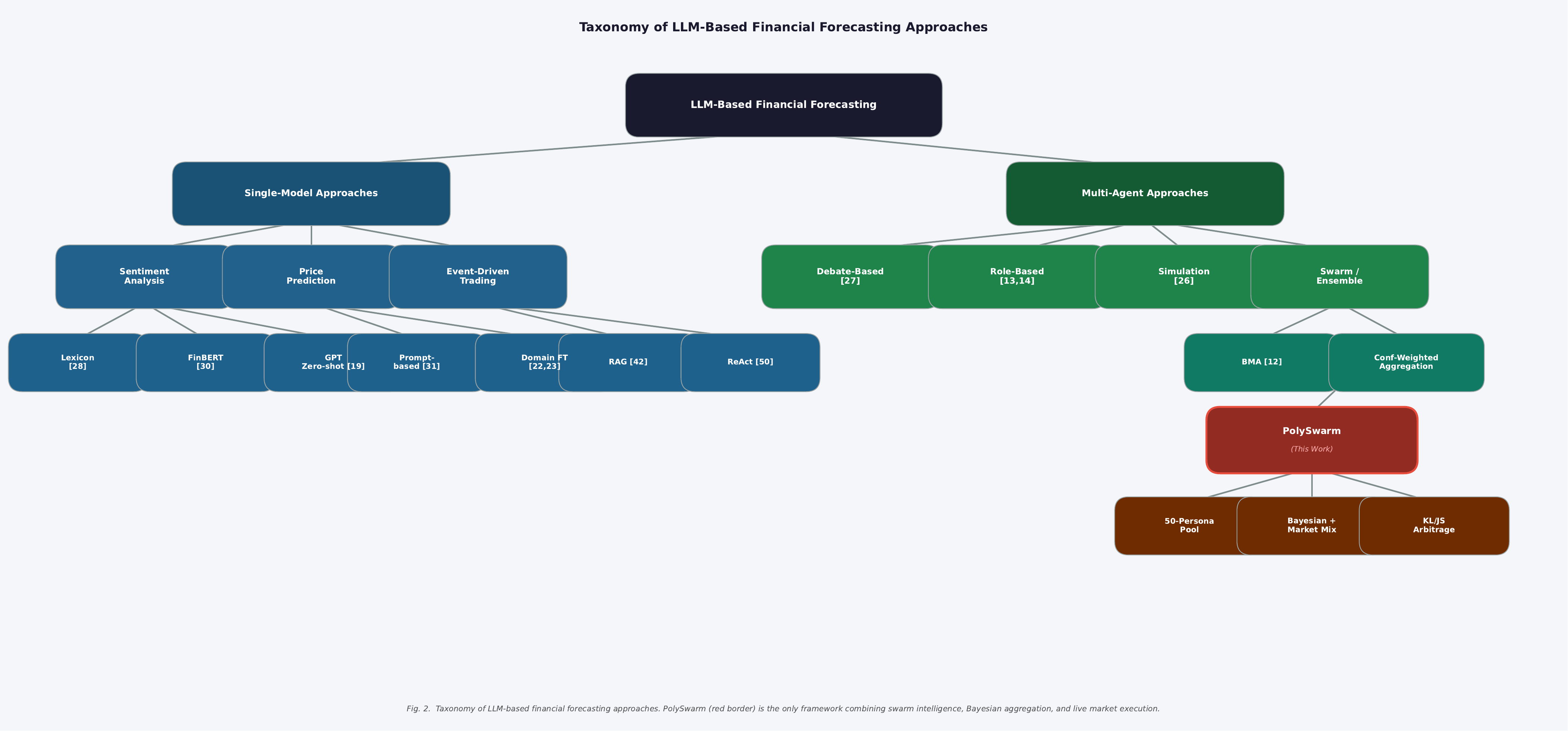}
  \caption{Taxonomy of LLM-based financial forecasting approaches.
  Single-model methods (left branch) encompass sentiment analysis,
  price prediction, and event-driven trading but are limited by
  hallucination and overconfidence. Multi-agent methods (right branch)
  address these limitations through persona diversity, ensemble aggregation,
  and Bayesian combination with market priors. PolySwarm instantiates the
  multi-agent swarm leaf of the taxonomy.}
  \label{fig:taxonomy}
\end{figure}

Figure~\ref{fig:taxonomy} situates PolySwarm within the broader landscape
of LLM-based financial forecasting approaches.
The remainder of the paper is organized as follows.
Section~\ref{sec:background} provides background on prediction markets,
large language models, and multi-agent systems.
Section~\ref{sec:llm_fin} reviews related work on LLMs for financial
forecasting.
Section~\ref{sec:architectures} presents the PolySwarm system architecture
and multi-agent design.
Section~\ref{sec:market_efficiency} develops the information-theoretic
framework for market analysis and arbitrage detection.
Section~\ref{sec:evaluation} describes evaluation methodology and benchmarks.
Section~\ref{sec:challenges} discusses open challenges, and
Section~\ref{sec:future} outlines future research directions.
Section~\ref{sec:conclusion} concludes.

\section{Background}
\label{sec:background}

\subsection{Prediction Markets}
\label{sec:pred_markets}

Prediction markets have a documented history extending back to the Iowa
Electronic Markets (IEM), launched in 1988 by the University of Iowa Tippie
College of Business as a research platform for studying collective
forecasting in political and economic domains.
The IEM demonstrated that market-aggregated probabilities could rival or
surpass traditional polling methodologies in forecasting election outcomes,
providing early empirical validation for the information-aggregation thesis
underlying prediction market design~\cite{wolfers2004prediction}.
The design principles governing information aggregation in combinatorial
prediction markets were subsequently formalized by
Hanson~\cite{hanson2003combinatorial}, whose logarithmic market scoring rule
(LMSR) provides a mechanism for subsidized market making that guarantees
bounded loss to the market operator while incentivizing honest probability
revelation by traders.
Subsequent platforms including InTrade, Hollywood Stock Exchange, and
PredictIt expanded the model to a broader range of outcome categories,
while academic research confirmed the general superiority of market-based
forecasts over expert panels and surveys across many
domains~\cite{arrow2008promise}.

Contemporary blockchain-based prediction markets represent the technological
frontier of the field.
Polymarket, launched in 2020 and operating on the Polygon
proof-of-stake network~\cite{nakamoto2008bitcoin,wood2014ethereum}, enables
permissionless participation in binary and categorical outcome markets with
on-chain settlement in USDC stablecoin.
By 2024, Polymarket had accumulated over \$500 million in trading volume,
including extensive activity around the United States presidential election,
Federal Reserve interest rate decisions, and cryptocurrency price milestones.
Kalshi, by contrast, operates as a CFTC-regulated designated contract market
in the United States, providing regulatory clarity at the cost of
jurisdictional restrictions on participation.
PredictIt, operated under a no-action letter from the CFTC for academic
research purposes, has served as an important dataset source for academic
studies of prediction market efficiency.

The theoretical foundations of prediction markets rest on the efficient
markets hypothesis (EMH) of Fama~\cite{fama1970efficient} and its
tension with the Grossman--Stiglitz paradox~\cite{grossman1980impossibility}.
Fama's weak-form EMH holds that asset prices fully reflect all publicly
available information, implying that systematic outperformance through
analysis of public data is impossible in equilibrium.
Grossman and Stiglitz identified a fundamental contradiction: if prices
are fully informative, rational agents have no incentive to incur the cost
of information acquisition, yet if no agent acquires information, prices
cannot be informative.
This paradox motivates the existence of informed traders who earn rents
proportional to their information advantage, sustaining the market mechanism
that produces price efficiency.
Multi-agent LLM systems occupy an interesting position in this framework:
they incur computational information-processing costs in exchange for
probability estimates that may differ from market consensus, and they trade
on those differences when the expected value of the discrepancy exceeds a
threshold.

\subsection{Large Language Models}
\label{sec:llms}

The transformer architecture introduced by Vaswani et
al.~\cite{vaswani2017attention} in 2017 constitutes the foundational
technical substrate for modern large language models.
The transformer's multi-head self-attention mechanism enables the model to
compute context-sensitive token representations across arbitrarily long
input sequences, overcoming the sequential bottleneck of recurrent neural
networks and enabling the scale of pretraining that characterizes
contemporary LLMs.
GPT-2~\cite{radford2019language}, released by OpenAI in 2019, first
demonstrated that a language model trained purely on next-token prediction
at scale could generate coherent long-form text and perform rudimentary
zero-shot task transfer, establishing the scaling hypothesis that motivated
subsequent generations of large models.
GPT-3~\cite{brown2020language}, released in 2020 with 175 billion
parameters, demonstrated remarkable few-shot generalization across a broad
range of natural language tasks without task-specific fine-tuning, marking
a qualitative inflection point in the capabilities of language models.
GPT-4~\cite{openai2023gpt4}, with an estimated parameter count in the
range of one trillion, extended these capabilities to multimodal inputs,
substantially longer context windows (up to 128K tokens), and demonstrably
improved reasoning and instruction-following.

Reinforcement learning from human feedback (RLHF), introduced for language
model alignment via InstructGPT by Ouyang et al.~\cite{ouyang2022training},
provides the fine-tuning methodology through which instruction-following
and safety properties are instilled in deployed chat models.
Open-source developments have substantially democratized access to
high-capability language models: the LLaMA family from Meta
AI~\cite{touvron2023llama} provides competitive performance across parameter
scales from 7B to 70B, enabling self-hosted deployment without dependence
on proprietary API providers.
LLaMA 3, released in 2024, extends context length to 8K tokens and delivers
substantially improved performance on reasoning benchmarks.

Domain-specific financial LLMs have emerged as a focused research direction.
BloombergGPT~\cite{wu2023bloomberggpt}, a 50-billion parameter model
pretrained on a curated corpus of 363 billion tokens of financial text
assembled by Bloomberg, demonstrates that domain-specific pretraining
yields substantial improvements on financial NLP benchmarks including
sentiment analysis, named entity recognition in financial documents, and
question answering over earnings transcripts.
FinGPT~\cite{yang2023fingpt} pursues an open-source, continuously updated
alternative by applying parameter-efficient fine-tuning techniques
(LoRA, QLoRA) to foundation models using streaming financial data,
offering a more accessible path to financial domain adaptation.
Chain-of-thought prompting~\cite{wei2022chain} has proven particularly
valuable in financial contexts, enabling models to produce explicit
intermediate reasoning steps that can be audited, improve accuracy on
multi-step quantitative reasoning, and reduce overconfident conclusions.
Self-consistency decoding~\cite{wang2023selfconsistency} further improves
reliability by sampling multiple reasoning chains and selecting the most
consistent answer.

Table~\ref{tab:llm_comparison} provides a comparison of prominent LLMs
relevant to financial applications along dimensions of organizational
provenance, scale, openness, financial domain fine-tuning, context capacity,
and reported performance on financial benchmarks.

\begin{table*}[!t]
  \renewcommand{\arraystretch}{1.3}
  \caption{Comparison of Prominent LLMs for Financial Applications}
  \label{tab:llm_comparison}
  \centering
  \begin{tabular}{lllcccl}
    \toprule
    \textbf{Model} & \textbf{Organization} & \textbf{Parameters} &
    \textbf{Open Source} & \textbf{Domain FT} & \textbf{Context} &
    \textbf{Fin.\ Benchmark} \\
    \midrule
    GPT-3         & OpenAI      & 175B      & No  & No  & 4K   & Moderate  \\
    GPT-4         & OpenAI      & $\sim$1T  & No  & No  & 128K & High      \\
    LLaMA 2       & Meta AI     & 7--70B    & Yes & No  & 4K   & Moderate  \\
    LLaMA 3       & Meta AI     & 8--70B    & Yes & No  & 8K   & High      \\
    BloombergGPT  & Bloomberg   & 50B       & No  & Yes & 2K   & Very High \\
    FinGPT        & Open        & 7--13B    & Yes & Yes & 4K   & High      \\
    Claude 3      & Anthropic   & N/A       & No  & No  & 200K & High      \\
    Mistral 7B    & Mistral AI  & 7B        & Yes & No  & 8K   & Moderate  \\
    \bottomrule
  \end{tabular}
\end{table*}

\subsection{Multi-Agent Systems}
\label{sec:mas}

Multi-agent systems (MAS) are computational architectures comprising
multiple autonomous agents that perceive their environment, pursue individual
or shared goals, and interact through communication or shared state.
In the classical AI and distributed computing literatures, MAS have been
applied to a wide range of problems including resource allocation, logistics
optimization, auction mechanisms, and robot coordination~\cite{dorri2018multi}.
The MAS paradigm is particularly well suited to problems characterized by
high-dimensional state spaces, distributed information, and requirements
for robust collective behaviour in the face of individual agent failures.

Swarm intelligence, as systematized by Bonabeau, Dorigo, and
Theraulaz~\cite{bonabeau1999swarm}, studies the collective intelligent
behaviour that emerges from the interaction of large numbers of relatively
simple agents without centralized control.
Classical swarm algorithms including Ant Colony Optimization and Particle
Swarm Optimization~\cite{kennedy1995particle} demonstrate that near-optimal
solutions to complex combinatorial and continuous optimization problems
can be found through decentralized stigmergic interaction.
The key design principles of swarm intelligence---agent diversity,
stochastic sampling, local interaction, and emergent global coordination---
translate directly to the design of LLM swarm systems.

Contemporary multi-agent LLM frameworks have operationalized these ideas
for language model orchestration.
AutoGen~\cite{wu2023autogen} provides a flexible conversation-based
framework in which LLM agents exchange structured messages, enabling
multi-turn collaborative problem solving, code generation, and tool use
across heterogeneous agent roles.
CAMEL~\cite{li2023camel} (Communicative Agents for Mind Exploration of
Large Language Models) employs a role-playing paradigm in which agents
adopt assigned social roles and collaboratively solve tasks through
structured dialogue, exploring the emergent collaborative behaviours of
LLM societies.
MetaGPT~\cite{hong2023metagpt} encodes software engineering workflows as
multi-agent collaboration graphs, assigning standardized roles (product
manager, architect, engineer, QA) to LLM agents and coordinating their
outputs through structured artefact handoffs, demonstrating that explicit
role specification substantially reduces hallucination in complex multi-step
reasoning tasks.
Generative Agents~\cite{park2023generative} demonstrated that LLM-based
agents equipped with persistent memory, reflection, and planning capabilities
can produce convincingly human-like social behaviour in a simulated
environment, opening research directions in agent architectures for
open-ended domains.
Multi-agent debate~\cite{liang2023encouraging} leverages the adversarial
dynamics of multiple agents arguing divergent positions to elicit more
accurate and well-calibrated conclusions than single-model self-consistency.
The wisdom-of-crowds literature~\cite{surowiecki2004wisdom} provides
theoretical grounding for why such ensemble approaches outperform individuals:
when individual errors are diverse and weakly correlated, aggregation
suppresses them, leaving signal intact.

\section{Related Work: LLMs for Financial Forecasting}
\label{sec:llm_fin}

\subsection{Sentiment Analysis}
\label{sec:sentiment}

Financial sentiment analysis has a long history predating the LLM era.
Early lexicon-based approaches assigned polarity scores to financial texts
by counting occurrences of positive and negative terms from curated
dictionaries, a methodology that achieved broad adoption in academic
research due to its interpretability and low computational cost.
Loughran and McDonald~\cite{loughran2011liability} made a seminal
contribution by demonstrating that general-purpose sentiment lexicons
such as the Harvard General Inquirer perform poorly in financial domains
because many terms with negative connotations in ordinary language
carry neutral or positive meaning in financial discourse
(e.g., ``liability,'' ``debt,'' ``tax'').
Their purpose-built financial sentiment dictionary has become a standard
baseline in the field.

The introduction of contextual word embeddings and transformer-based
language models fundamentally changed the capabilities available to
financial sentiment analysis.
BERT~\cite{devlin2019bert}, by producing deeply bidirectional contextual
token representations, enabled classification of financial texts at a
level of semantic nuance inaccessible to lexicon methods or shallower
embedding models.
FinBERT~\cite{araci2019finbert}, a BERT model fine-tuned on a corpus of
financial news and analyst reports, demonstrated clear improvements over
general-purpose BERT on financial sentiment benchmarks, establishing
domain-specific fine-tuning as best practice for financial NLP.
Subsequent work has extended this approach to larger architectures:
GPT-based models applied to financial sentiment have demonstrated
significant improvements on benchmarks including Financial PhraseBank,
FiQA, and earnings call sentiment classification~\cite{lopezlira2023chatgpt}.

Lopez-Lira and Tang~\cite{lopezlira2023chatgpt} published an influential
study demonstrating that ChatGPT-generated sentiment scores for news
headlines exhibit predictive power for next-day stock returns that is
substantially superior to lexicon-based approaches, survives standard
risk-factor controls, and displays statistically significant alpha.
This finding has catalyzed a wave of research examining whether LLMs can
serve as effective zero-shot financial analysts, bypassing the expensive
supervised training pipelines that characterize previous generation
approaches.

\subsection{Price Prediction}
\label{sec:price_pred}

Direct application of LLMs to price prediction represents a more
ambitious and contested research direction.
Zero-shot and few-shot approaches prompt LLMs with recent price histories,
relevant news, and technical indicators in natural language form, soliciting
directional predictions (up/down/neutral) or explicit probability estimates
for future price movements.
Early demonstrations showed promising results on specific equities and time
periods~\cite{lopezlira2023chatgpt}, but subsequent rigorous evaluations
have revealed important limitations.
LLMs lack access to real-time market data and frequently exhibit
miscalibrated confidence in predictions, particularly for high-volatility
events that by their nature involve distributional tail behaviour
underrepresented in training corpora.
The strong version of the EMH implies that consistently profitable
price prediction from publicly available textual information should be
impossible in liquid markets, suggesting that LLM alpha in this domain
may be concentrated in less efficient market segments, may decay as
market participants adapt to LLM signals, or may reflect
overfitting in evaluation protocols.

Few-shot prompting approaches that provide LLMs with carefully curated
examples of successful analyses alongside the target query have shown
modest improvements over zero-shot baselines, but require careful
example selection to avoid inadvertent look-ahead bias.
Retrieval-augmented generation (RAG)~\cite{lewis2020retrieval} offers a
promising extension by enabling models to dynamically retrieve relevant
historical precedents and current context from external knowledge stores,
partially addressing the staleness of static training data.
However, the coherent integration of retrieved information with parametric
knowledge remains an active research challenge, particularly when retrieved
documents conflict with the model's prior beliefs.

\subsection{News and Event-Driven Trading}
\label{sec:news_trading}

News and event-driven trading represents perhaps the most natural application
of LLMs to financial markets, given that language models are explicitly
trained on the textual data formats that news and corporate disclosures
take.
The core pipeline involves ingesting a continuous stream of breaking news
and announcements, classifying their relevance and directional impact on
specific securities or prediction market contracts, and generating trading
signals faster than human analysts can process the same information.
LLMs offer advantages over earlier natural language processing approaches
in this pipeline through their ability to comprehend nuanced, context-dependent
language, resolve ambiguous coreferences, and reason about the downstream
implications of events for market prices without explicit training on
event-specific examples.

Retrieval-augmented generation~\cite{lewis2020retrieval} substantially
enhances event-driven LLM systems by equipping them with dynamic access
to background context including company filings, historical price reactions
to similar events, macroeconomic data, and prior news coverage.
This retrieval layer compensates for the static nature of pretrained
parametric knowledge and enables the model to contextualize breaking
information against relevant historical precedent.
The ReAct framework~\cite{yao2023react}, which interleaves reasoning traces
with action execution (including information retrieval and external tool
calls), has demonstrated particular promise for financial analysis agents
that must gather, integrate, and act on information across multiple
heterogeneous sources.
Self-refinement approaches such as Reflexion~\cite{shinn2023reflexion}
enable agents to learn from their prediction errors within a session,
iteratively improving their analysis through explicit verbal reflection
on prior mistakes.

\subsection{Limitations of Single-Model Approaches}
\label{sec:single_model_lim}

The financial application of LLMs is constrained by a set of systematic
limitations that apply with particular force when models are deployed in
isolation rather than as components of a larger ensemble architecture.
Hallucination---the generation of confident, fluent, but factually
incorrect outputs~\cite{ji2023hallucination}---poses an acute risk in
financial contexts where erroneous factual claims about company performance,
regulatory status, or macroeconomic indicators could drive incorrect
trading decisions.
Unlike natural language tasks where hallucinations may simply produce
unhelpful outputs, financial hallucinations can result in direct monetary
losses, and their confident presentation makes them especially dangerous
in automated pipelines that lack human review at inference time.

Overconfidence and miscalibration represent a related but distinct failure
mode~\cite{kadavath2022language}.
LLMs trained to produce helpful, decisive responses tend to generate
high-confidence probability estimates even in domains characterized by
genuine fundamental uncertainty, producing poorly calibrated outputs that
systematically overstate the model's epistemic certainty.
In a prediction market context, overconfidence translates directly to
excessive position sizing and exposure to tail-risk events.
Prompt sensitivity---the phenomenon whereby superficially equivalent
phrasings of the same query produce substantially different
outputs~\cite{zhao2021calibrate}---undermines the reproducibility and
reliability of single-model financial analysis, since the particular
prompt template chosen by a system designer may inadvertently anchor
model outputs in non-representative directions.
Finally, and most fundamentally, a single LLM instance represents a
single draw from a high-variance inference distribution.
Without epistemic diversity across the agent pool, correlated errors
cannot be averaged away, and the system inherits all the biases and
blind spots of its single analytical perspective~\cite{surowiecki2004wisdom}.

Table~\ref{tab:literature_summary} provides a summary of representative
LLM-based financial forecasting studies drawn from the literature reviewed
in this section, together with the PolySwarm system presented in this paper.

\begin{table*}[!t]
  \renewcommand{\arraystretch}{1.3}
  \caption{Summary of Representative LLM-Based Financial Forecasting Studies}
  \label{tab:literature_summary}
  \centering
  \begin{tabular}{p{2.6cm} c p{2.2cm} p{2.2cm} p{3.0cm} p{4.0cm}}
    \toprule
    \textbf{Reference} & \textbf{Year} & \textbf{Model} &
    \textbf{Task} & \textbf{Method} & \textbf{Key Result} \\
    \midrule
    Lopez-Lira \& Tang~\cite{lopezlira2023chatgpt}
      & 2023 & ChatGPT & Stock ret.\ prediction
      & Zero-shot headline sentiment
      & Significant alpha over lexicon baselines \\
    \addlinespace
    Wu et al.~\cite{wu2023bloomberggpt}
      & 2023 & BloombergGPT & Multiple fin.\ NLP tasks
      & Domain-specific pretraining
      & SOTA on financial benchmarks \\
    \addlinespace
    Yang et al.~\cite{yang2023fingpt}
      & 2023 & FinGPT & Sentiment, Q\&A
      & RLHF fine-tuning on streaming data
      & Competitive with proprietary models \\
    \addlinespace
    Araci~\cite{araci2019finbert}
      & 2019 & FinBERT & Financial sentiment
      & Domain-adapted BERT
      & Outperforms general BERT on PhraseBank \\
    \addlinespace
    Liang et al.~\cite{liang2023encouraging}
      & 2023 & GPT-4 & Reasoning accuracy
      & Multi-agent debate
      & Reduced hallucination vs.\ single model \\
    \addlinespace
    Wu et al.~\cite{wu2023autogen}
      & 2023 & GPT-4 & Multi-step reasoning
      & Multi-agent conversation (AutoGen)
      & Flexible agent orchestration \\
    \addlinespace
    Lewis et al.~\cite{lewis2020retrieval}
      & 2020 & BERT + RAG & Open-domain QA
      & Retrieval-augmented generation
      & Strong performance on knowledge-intensive tasks \\
    \addlinespace
    Yao et al.~\cite{yao2023react}
      & 2023 & GPT-3 & Decision-making tasks
      & ReAct (reasoning + acting)
      & Outperforms CoT alone on tool-use tasks \\
    \addlinespace
    Wei et al.~\cite{wei2022chain}
      & 2022 & GPT-3/4 & Mathematical reasoning
      & Chain-of-thought prompting
      & Large gains on multi-step reasoning \\
    \addlinespace
    Park et al.~\cite{park2023generative}
      & 2023 & GPT-3.5 & Social simulation
      & Generative agent architecture
      & Human-like emergent social behaviour \\
    \addlinespace
    \textbf{PolySwarm (This work)}
      & 2026 & Multi-provider LLMs & Prediction market trading
      & 50-persona swarm + Bayesian aggregation + KL divergence
      & End-to-end paper/live trading with arbitrage detection \\
    \bottomrule
  \end{tabular}
\end{table*}

\section{PolySwarm System Design and Architecture}
\label{sec:architectures}

\subsection{Agent Design Patterns}
\label{sec:agent_design}

The design of individual agents within a multi-agent LLM system requires
decisions across three primary dimensions: persona specification, memory
architecture, and information access patterns.
Persona design determines the analytical perspective, prior beliefs, and
reasoning style that a given agent will exhibit.
In homogeneous multi-agent systems, all agents receive identical prompts
and differ only through stochastic sampling; while this approach is simple
to implement, it limits the epistemic diversity that is essential for
effective ensemble aggregation.
Heterogeneous persona design, by contrast, assigns distinct roles,
backgrounds, and analytical frameworks to different agents in the pool,
deliberately engineering the diversity that the wisdom-of-crowds effect
requires~\cite{surowiecki2004wisdom}.
Representative persona archetypes for financial forecasting include
momentum traders (who weight recent price trend signals), contrarian
analysts (who specifically seek out consensus errors), macro economists
(who anchor analysis in aggregate indicators), technical analysts
(who reason from chart patterns and volume signals), and fundamental
investors (who focus on earnings growth and balance sheet quality).
Each archetype brings a systematically different analytical lens, ensuring
that the aggregate output of the swarm reflects a genuinely multi-dimensional
evaluation of the market in question.

Memory architecture governs how agents access, update, and retrieve
information across turns and sessions.
Stateless agents, which process each query independently without reference
to prior session history, are simplest to implement but cannot learn from
or adapt to feedback on previous predictions.
Short-term memory, implemented via extended prompt context windows that
include recent conversation history, enables agents to maintain analytical
coherence across multi-step reasoning chains.
Long-term memory, implemented via external vector databases or structured
knowledge stores with embedding-based retrieval, enables agents to recall
relevant historical precedents and prior analyses that lie outside the
current context window.
LLM caching, which stores the responses to previously seen queries and
serves cached responses for similar future queries within a configurable
time-to-live window, reduces both computational latency and API cost in
high-throughput deployments---an important practical consideration for
systems making hundreds of LLM calls per scan cycle.

Information access patterns determine what data each agent can observe
when formulating its prediction.
Agents may be given access to current market prices, recent volume data,
breaking news feeds, macroeconomic data releases, social media sentiment
aggregates, historical resolution data for similar markets, and the
predictions of other agents in the swarm.
The design choice of whether to provide agents with the current market
probability before eliciting their prediction is particularly consequential:
anchoring agents to the market price risks suppressing genuine disagreement,
but withholding it may cause agents to ignore a strong informative prior.
In PolySwarm, market-implied probabilities are withheld during individual
agent inference and incorporated only at the Bayesian aggregation stage,
preserving independence of individual agent predictions while still
leveraging market information in the final combined estimate.

\subsection{Communication and Coordination}
\label{sec:communication}

The manner in which agents communicate and coordinate their analyses
within a multi-agent system has profound implications for the quality
and diversity of collective output.
The simplest coordination regime is \emph{independent parallel sampling}:
agents receive identical or persona-differentiated versions of the same
query, produce independent predictions, and submit results to a central
aggregator without any inter-agent communication.
This approach maximizes prediction diversity because agents cannot anchor
to or be influenced by each other's outputs, preserving the statistical
independence that enables error cancellation in ensemble aggregation.
The principal disadvantage is that independent agents cannot pool information
or resolve factual disputes, and may produce systematically divergent
analyses based on inconsistent readings of ambiguous evidence.

\emph{Multi-agent debate}~\cite{liang2023encouraging} offers an alternative
coordination regime in which agents first produce independent initial
analyses, then observe each other's reasoning, and iteratively revise
their positions through structured argumentation rounds.
This dialectical process encourages agents to surface and scrutinize
implicit assumptions, challenge poorly supported claims, and converge
on more accurate and better-calibrated conclusions than independent
sampling alone.
Empirical evaluations have demonstrated that multi-agent debate reduces
factual hallucination rates and improves reasoning accuracy across a range
of benchmarks, at the cost of substantially increased computational overhead
from multiple inference rounds.

\emph{Blackboard architectures} centralize inter-agent communication
through a shared global data structure (the ``blackboard'') to which
agents post observations and from which they read the contributions of
others.
This architecture supports asynchronous coordination and is well suited
to systems in which agents have heterogeneous information access and
processing speeds.
In financial applications, the blackboard might contain a continuously
updated summary of market data, news events, and prior agent analyses,
enabling late-arriving agents to integrate prior work without waiting
for synchronous turn-taking.

\emph{Voting and averaging} methods aggregate agent predictions through
simple majority voting (for categorical outputs), arithmetic mean
probability estimation, or more sophisticated ensemble methods including
confidence-weighted averaging and Bayesian model averaging.
The choice among these methods involves a trade-off between simplicity,
robustness to outliers, and exploitation of differential agent reliability
information.
In well-calibrated agent pools where all agents are roughly equally
reliable, simple arithmetic averaging performs competitively with more
sophisticated methods; when agent quality is heterogeneous, confidence
or performance weighting yields material improvements.

\subsection{Bayesian Aggregation}
\label{sec:bayes_agg}

Bayesian model averaging (BMA)~\cite{hoeting1999bayesian} provides the
principled statistical framework for combining predictions from multiple
models or agents.
In its canonical form, BMA computes the combined predictive distribution
as a weighted average of individual model predictive distributions, with
weights proportional to the posterior probability of each model given
the observed data.
In the LLM swarm context, the posterior model weights are not directly
computable from first principles, but can be approximated using
track-record-based estimates of agent reliability, calibration scores,
or domain-specific performance metrics.

A two-stage Bayesian aggregation procedure is natural for prediction market
forecasting systems that wish to combine swarm consensus with market-implied
probability.
In the first stage, individual agent predictions are aggregated into a
swarm consensus probability $p_{\text{swarm}}$ through confidence-weighted
averaging:
\begin{equation}
  p_{\text{swarm}} = \frac{\sum_{i=1}^{N} w_i \, p_i}{\sum_{i=1}^{N} w_i},
\end{equation}
where $p_i$ is agent $i$'s predicted probability, $w_i$ is agent $i$'s
confidence or reliability weight, and $N$ is the number of agents in the
pool.
In the second stage, the swarm consensus is combined with the market-implied
probability $p_{\text{market}}$ through a linear Bayesian mixture:
\begin{equation}
  p_{\text{combined}} = 0.70 \times p_{\text{swarm}} + 0.30 \times p_{\text{market}}.
  \label{eq:bayesian_combo}
\end{equation}
The 70/30 weighting in Eq.~(\ref{eq:bayesian_combo}) encodes a prior
belief that the swarm's independent analysis should dominate the final
estimate while still incorporating the substantial information content
of the market price.
This weighting is a tunable hyperparameter; increasing the market weight
causes the system to behave more conservatively and trade less frequently,
while increasing the swarm weight allows the system to take larger
positions on the basis of its own analysis relative to the market consensus.
Gneiting and Raftery~\cite{gneiting2007strictly} establish that
proper scoring rules such as the Brier score and logarithmic score provide
the correct incentives for agents to report their true predictive
probabilities, making them the appropriate loss functions for calibrating
individual agent weights.

\subsection{Swarm-Based Approaches: The PolySwarm System}
\label{sec:polyswarm}

PolySwarm is a production multi-agent swarm trading terminal that implements
the architectural principles described in the preceding subsections in the
context of live Polymarket prediction market trading.
The system is built on a Python FastAPI backend with a Vue 3 frontend
communicating via WebSocket for real-time dashboard updates.

\textbf{Agent pool.}
PolySwarm maintains a pool of 50 diverse LLM personas (the \texttt{PERSONA\_POOL}),
each defined by a distinct analytical archetype, personality profile, and
reasoning orientation.
Persona archetypes include macro economists, technical analysts, contrarian
investors, political scientists, sports statisticians, public health experts,
and domain specialists across the full range of prediction market categories.
For each market evaluation, a configurable number of agents (default: 25)
is sampled from the pool without replacement, ensuring that successive
evaluations draw on varied analytical perspectives while keeping per-scan
computational costs bounded.
Each agent formulates its prediction via a structured chain-of-thought
prompt~\cite{wei2022chain} that requires explicit articulation of supporting
reasoning, uncertainty sources, and confidence level before committing to
a numerical probability estimate.
This chain-of-thought elicitation improves calibration by forcing agents
to confront the considerations that bear on their estimate, reduces
overconfident snap judgements, and produces audit trails that can be
reviewed for quality assurance.

\textbf{Scan loop and concurrency.}
The core scan loop executes on a 5-second cycle, ingesting active markets
from Polymarket's Gamma REST API, filtering by minimum trading volume and
recent activity thresholds, and dispatching swarm evaluations for markets
that meet selection criteria.
Concurrent LLM inference is managed through Python's \texttt{asyncio}
framework with a bounded semaphore controlling maximum simultaneous
in-flight API requests, providing rate-limit compliance and predictable
latency under variable market load.
Multi-provider support enables the system to distribute inference across
Anthropic (Claude series), OpenAI (GPT series), and self-hosted Ollama
(LLaMA, Mistral) backends, allowing cost-quality trade-offs to be tuned
per deployment context.
LLM responses are cached in an asynchronous SQLite database with a
configurable time-to-live, avoiding redundant API calls for markets whose
information state has not materially changed since the last evaluation.

\textbf{Aggregation and expected value.}
After all sampled agents have returned predictions, the first-stage
confidence-weighted average produces $p_{\text{swarm}}$, and the second-stage
Bayesian combination produces $p_{\text{combined}}$ per Eq.~(\ref{eq:bayesian_combo}).
The expected value (EV) of a candidate trade is computed as:
\begin{equation}
  \text{EV} = p_{\text{combined}} \times b - (1 - p_{\text{combined}}),
\end{equation}
where $b$ is the net decimal odds of the YES outcome implied by the current
market price.
Trades are triggered only when EV exceeds a configurable minimum threshold
(default: 5\%) and swarm standard deviation is below 30\%, the latter
condition preventing position entry when swarm consensus is wide and
epistemic uncertainty is high.

\textbf{Trading execution.}
In paper trading mode (the default), orders are simulated in memory with
full tracking of virtual positions, PnL, and win rate.
In live trading mode, orders are submitted to Polymarket's CLOB (Central
Limit Order Book) API via the \texttt{py-clob-client} library, executing
real transactions on the Polygon blockchain.
Position sizing follows the quarter-Kelly criterion (Section~\ref{sec:position_sizing}),
with a hard maximum position cap (\texttt{MAX\_POSITION\_USDC}) enforced
at the execution layer.
A daily loss limit (\texttt{DAILY\_LOSS\_LIMIT\_USDC}) automatically
suspends the scan loop when hit, providing an essential first-order
risk management safeguard.

\begin{figure}[!t]
  \centering
  \includegraphics[width=\columnwidth]{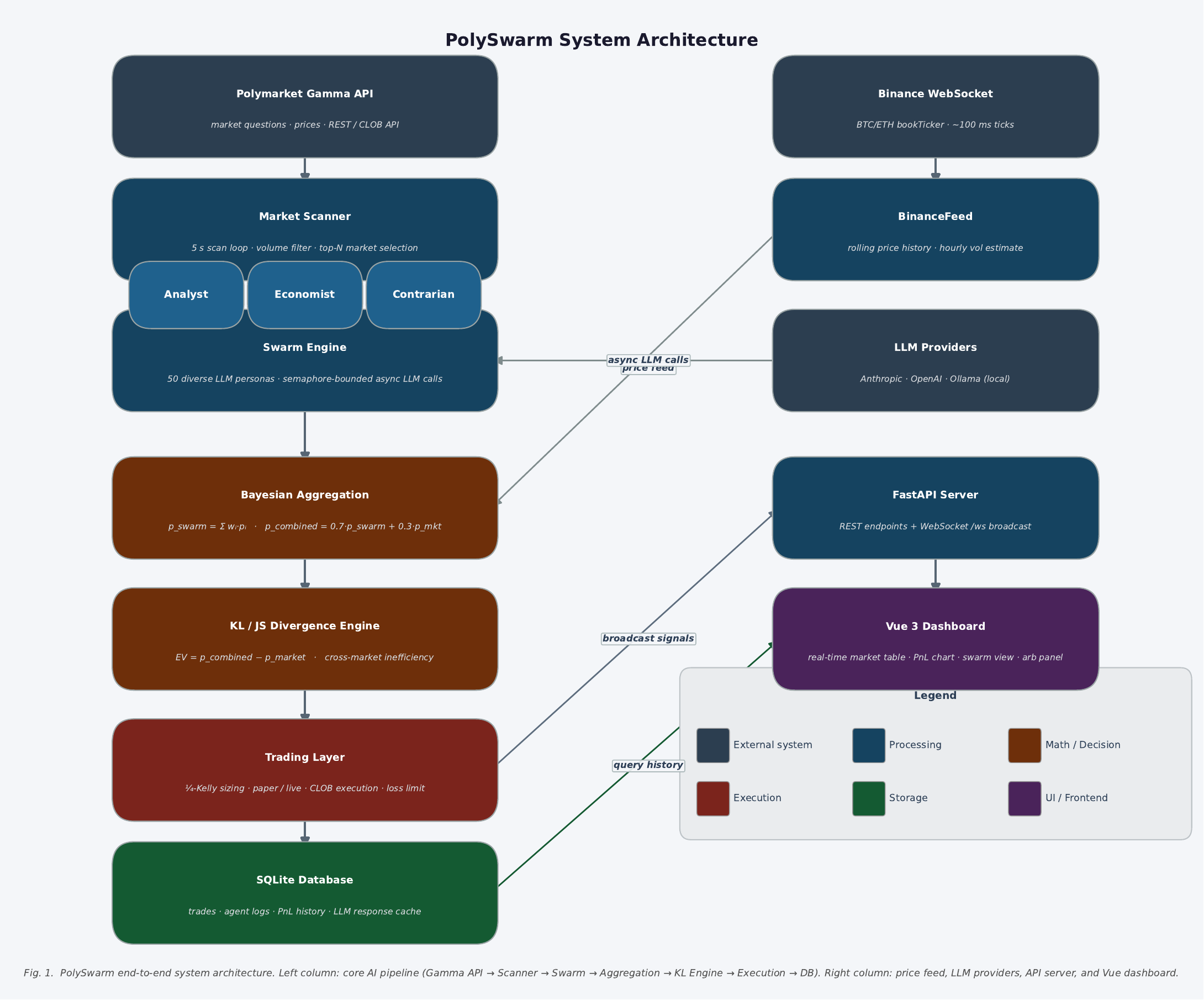}
  \caption{PolySwarm system architecture. Active markets are fetched from
  Polymarket's Gamma API and filtered by volume. The swarm engine samples
  $N$ agents from the 50-persona pool and fires concurrent LLM inference
  calls with a rate-limiting semaphore. Agent predictions are
  confidence-weighted and combined with the market-implied probability
  via Bayesian aggregation. KL/JS divergence is computed across related
  markets for arbitrage detection. Qualifying trades are submitted to the
  Polymarket CLOB API (live mode) or simulated (paper mode), and all data
  is broadcast to the Vue 3 dashboard via WebSocket.}
  \label{fig:architecture}
\end{figure}

\textbf{Database and monitoring.}
All market snapshots, individual agent predictions, trade executions,
detected inefficiencies, and daily PnL records are persisted in an
asynchronous SQLite database via \texttt{aiosqlite}.
A FastAPI REST and WebSocket server exposes this data to a real-time
Vue 3 dashboard comprising market tables, probability charts, PnL history,
agent swarm visualizations, and a live log feed.
Figure~\ref{fig:architecture} presents the complete end-to-end system
architecture.

\subsection{Comparison of Architectures}
\label{sec:arch_comparison}

Table~\ref{tab:framework_comparison} compares PolySwarm against the
representative multi-agent LLM frameworks reviewed in this section along
the dimensions most relevant to financial forecasting applications.

\begin{table*}[!t]
  \renewcommand{\arraystretch}{1.3}
  \caption{Comparison of Multi-Agent LLM Frameworks}
  \label{tab:framework_comparison}
  \centering
  \begin{tabular}{lcccccc}
    \toprule
    \textbf{Framework} & \textbf{Agent Count} & \textbf{Coordination} &
    \textbf{Aggregation} & \textbf{Financial Focus} &
    \textbf{Real-Time} & \textbf{Open Source} \\
    \midrule
    AutoGen~\cite{wu2023autogen}
      & 2--10     & Conversational   & Consensus dialogue  & No  & No  & Yes \\
    CAMEL~\cite{li2023camel}
      & 2--5      & Role-play        & Emergent agreement  & No  & No  & Yes \\
    AgentVerse~\cite{chen2023agentverse}
      & Variable  & Dynamic grouping & Voting              & No  & No  & Yes \\
    Multi-Agent Debate~\cite{liang2023encouraging}
      & 2--6      & Debate rounds    & Final round output  & No  & No  & Yes \\
    Generative Agents~\cite{park2023generative}
      & 25        & Social sim.      & Emergent behaviour  & No  & No  & Yes \\
    \textbf{PolySwarm (This work)}
      & \textbf{50} & \textbf{Independent parallel} &
      \textbf{Bayesian weighted} & \textbf{Yes} & \textbf{Yes} & \textbf{Yes} \\
    \bottomrule
  \end{tabular}
\end{table*}

\section{Market Efficiency Analysis and Arbitrage Detection}
\label{sec:market_efficiency}

\subsection{Information-Theoretic Approaches}
\label{sec:info_theory}

Information theory provides a rigorous mathematical framework for quantifying
the divergence between probability distributions, making it a natural
analytical tool for comparing swarm-estimated and market-implied probabilities
in prediction markets.
The Kullback--Leibler (KL) divergence~\cite{kullback1951information},
also known as relative entropy, measures the information gain achieved
by revising from distribution $Q$ to distribution $P$:
\begin{equation}
  D_{\mathrm{KL}}(P \,\|\, Q) = \sum_{x} P(x) \log \frac{P(x)}{Q(x)},
  \label{eq:kl_divergence}
\end{equation}
where the sum is taken over all outcomes $x$ in the shared support of
$P$ and $Q$.
In the prediction market context, $P$ represents the swarm consensus
distribution and $Q$ represents the market-implied distribution.
A large KL divergence value in Eq.~(\ref{eq:kl_divergence}) indicates that
the swarm's probability estimate departs substantially from the market
consensus, signalling a potential market inefficiency that the system may
wish to investigate and potentially trade.

The Jensen--Shannon (JS) divergence~\cite{cover2006elements} addresses
the asymmetry of KL divergence (since $D_{\mathrm{KL}}(P\|Q) \neq
D_{\mathrm{KL}}(Q\|P)$ in general) by constructing a symmetric and bounded
alternative:
\begin{equation}
  D_{\mathrm{JS}}(P \,\|\, Q) = \frac{1}{2} D_{\mathrm{KL}}(P \,\|\, M)
                               + \frac{1}{2} D_{\mathrm{KL}}(Q \,\|\, M),
\end{equation}
where $M = \frac{1}{2}(P + Q)$ is the mixture distribution.
The JS divergence is bounded in $[0, \log 2]$ (for natural logarithms),
making it easier to interpret as a normalized measure of distributional
distance.
Its square root, the Jensen--Shannon distance, satisfies the triangle
inequality and defines a proper metric on the space of probability
distributions.
PolySwarm computes both KL and JS divergence for each evaluated market
as part of its inefficiency scoring pipeline, using them as primary inputs
to the decision of whether to flag a market for trading consideration.
Markets with high JS divergence between swarm and market distributions
are prioritized for position entry, subject to EV and uncertainty filters.

\subsection{Cross-Market Inefficiency Detection}
\label{sec:cross_market}

Beyond single-market divergence, prediction market ecosystems frequently
exhibit cross-market inconsistencies that reveal structural mispricings
absent from any single contract.
\emph{Negation pairs} are among the most tractable of these: if a market
offers contracts on event $E$ and a separate market offers contracts on
$\neg E$ (the logical negation of $E$), the no-arbitrage condition requires
that $P(E) + P(\neg E) = 1$.
Deviations from this condition, which can arise from differential liquidity,
participant demographics, or market timing mismatches, represent a direct
arbitrage opportunity: simultaneously buying the underpriced side and selling
(or not buying) the overpriced side locks in a risk-free gain at resolution.
PolySwarm's cross-market analysis module identifies negation pairs using
semantic similarity matching over market title strings, computes the implied
probability sum, and flags pairs where the sum deviates from unity by more
than a configurable threshold.

More complex cross-market constraints arise from \emph{mutually exclusive
outcome markets} and \emph{Bayesian network consistency} conditions.
When multiple prediction markets collectively cover an exhaustive,
mutually exclusive partition of outcomes (e.g., Q1, Q2, Q3, Q4 economic
growth categories), the probabilities assigned to each partition must
sum to one.
Violations of this sum constraint indicate that the market as a whole
is mispriced, and optimal trading involves taking positions in the
underpriced outcomes and avoiding the overpriced ones.
Bayesian network analysis can detect more subtle consistency violations
among correlated markets---for example, if the market-implied conditional
probability of outcome $B$ given $A$ is inconsistent with the unconditional
probabilities of $A$ and $B$---providing additional signals of structural
inefficiency.

\subsection{Latency Arbitrage}
\label{sec:latency_arb}

Latency arbitrage exploits the temporal lag between the arrival of
price-relevant information and its incorporation into market prices.
In traditional equity markets, the high-frequency trading (HFT) arms
race~\cite{budish2015hft} has reduced exploitable latency gaps to
microsecond timescales through co-location, direct market access, and
microwave communication networks~\cite{menkveld2013hft}.
The blockchain-based prediction market context presents a qualitatively
different latency landscape.
Smart contract execution and on-chain settlement introduce deterministic
block-time latencies (12 seconds for Ethereum mainnet, 2 seconds for
Polygon PoS), and the oracle update cycles that feed external event
resolutions to prediction markets may lag the real-world event by
minutes to hours.
For cryptocurrency price contracts, PolySwarm's latency arbitrage engine
derives a CEX-implied probability using the log-normal price model
underlying the Black--Scholes framework~\cite{black1973pricing},
computing $p_{\mathrm{cex}} = \Phi\!\left(\ln(S/K)/(\sigma\sqrt{T})\right)$
where $S$ is the current spot price, $K$ the contract strike, $\sigma$ the
hourly volatility, and $T$ the time to expiry in hours.
Divergence between $p_{\mathrm{cex}}$ and the stale Polymarket price $p_{\mathrm{poly}}$
signals an exploitable edge.
Miner Extractable Value (MEV) in DeFi contexts~\cite{daian2020flash}
represents a form of blockchain-native latency arbitrage in which
miners or validators reorder, insert, or censor transactions to capture
value, a phenomenon with potential implications for prediction market
execution fairness.

\begin{figure}[!t]
  \centering
  \includegraphics[width=\columnwidth]{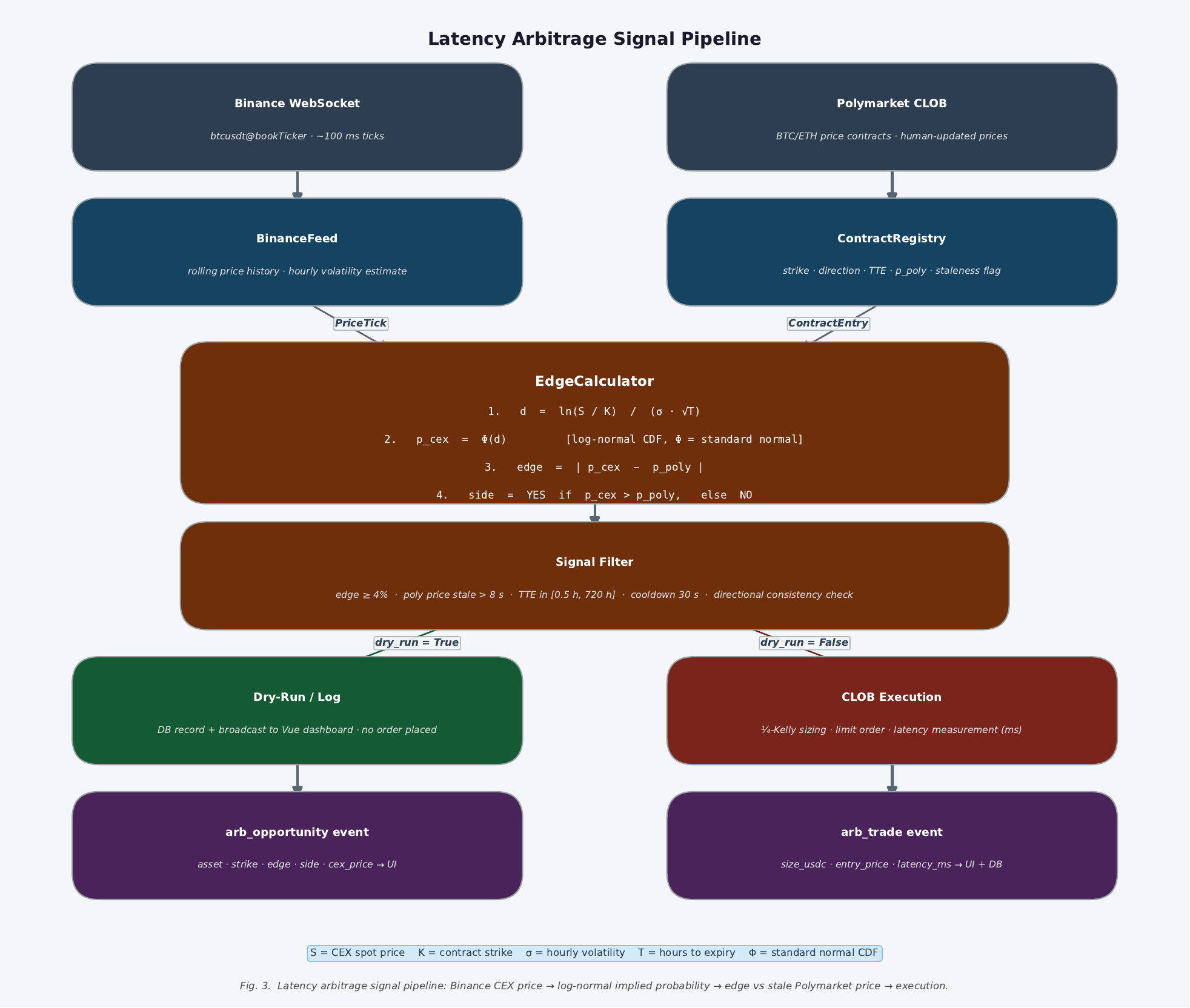}
  \caption{Latency arbitrage pipeline in PolySwarm. Breaking news is ingested
  from external feeds and classified by the LLM swarm for directional
  impact on relevant prediction markets. Market prices are polled on a
  5-second cycle. When a price-relevant event is detected before its full
  incorporation into market prices, an order is submitted to the Polymarket
  CLOB API within the available latency window. Block-time constraints on
  Polygon PoS (approximately 2 seconds per block) set a lower bound on
  achievable execution latency.}
  \label{fig:latency_arb}
\end{figure}

PolySwarm's 5-second scan loop is designed to operate at the practical
resolution limit of Polymarket's REST API, enabling the system to detect
price movements that lag real-world information arrivals.
Breaking news events, policy announcements, and election results frequently
require several minutes to be fully incorporated into prediction market
prices, as human traders read, process, and manually submit orders.
An automated system that can classify the directional implications of
breaking news and submit an order within seconds of publication exploits
this human processing lag.
The latency arbitrage pipeline integrates breaking news ingestion, LLM
classification, EV calculation, and order submission in a single asynchronous
pipeline designed to minimize end-to-end processing time, as illustrated in
Figure~\ref{fig:latency_arb}.

\subsection{Position Sizing: The Kelly Criterion}
\label{sec:position_sizing}

The Kelly criterion~\cite{kelly1956information} provides an information-theoretically
optimal position sizing formula derived from the objective of maximizing
the long-run expected logarithm of wealth.
For a binary bet with net odds $b$ (i.e., a winning stake of one unit
returns $b$ units profit), the Kelly fraction $f^*$ is:
\begin{equation}
  f^* = \frac{p \cdot b - (1-p)}{b},
  \label{eq:kelly}
\end{equation}
where $p$ is the estimated probability of the favourable outcome.
Equation~(\ref{eq:kelly}) specifies what fraction of bankroll to wager:
positive values indicate a favourable bet, and the formula automatically
scales position size with edge, allocating more capital when the edge is
large and less when it is small.

Full Kelly sizing is known to produce high portfolio volatility and extended
drawdown periods that are psychologically and practically challenging to
sustain~\cite{thorp2006kelly}.
The quarter-Kelly convention ($f = 0.25 \cdot f^*$), implemented in
PolySwarm via the \texttt{KELLY\_FRACTION} configuration parameter,
reduces variance substantially while retaining the majority of the
long-run growth-rate advantage.
The Kelly formula's dependence on the probability estimate $p$ means
that position sizing inherits the calibration properties of the underlying
prediction system: well-calibrated predictions produce appropriate position
sizes, while overconfident predictions produce excessive leverage.
This coupling motivates the uncertainty filter in PolySwarm (no trading
when swarm standard deviation exceeds 30\%), which prevents the system
from taking large positions when the swarm's own disagreement indicates
low confidence.
A hard maximum position cap (\texttt{MAX\_POSITION\_USDC}, default: \$10)
provides an absolute upper bound on per-trade exposure independent of
the Kelly calculation, ensuring that estimation errors in $p$ or $b$
cannot produce catastrophic single-trade losses.

\section{Evaluation Methodology}
\label{sec:evaluation}

\subsection{Forecasting Accuracy Metrics}
\label{sec:metrics}

Rigorous evaluation of probabilistic forecasting systems requires metrics
that reward calibration as well as discrimination, since a system that
consistently assigns extreme probabilities to resolved events may score
well on discrimination metrics while being poorly calibrated and thus
generating badly sized positions.
Tetlock and Gardner's work on superforecasters~\cite{tetlock2015superforecasting}
established that disciplined probabilistic thinkers can systematically
outperform both expert consensus and prediction market prices over large
question sets, providing a human performance benchmark against which
automated LLM systems should be measured.
The Brier score~\cite{brier1950verification}, the most widely used proper
scoring rule for binary probabilistic forecasts, is defined as:
\begin{equation}
  \text{BS} = \frac{1}{N} \sum_{t=1}^{N} (f_t - o_t)^2,
  \label{eq:brier}
\end{equation}
where $f_t \in [0,1]$ is the forecast probability for event $t$ and
$o_t \in \{0, 1\}$ is the binary outcome.
The Brier score in Eq.~(\ref{eq:brier}) ranges from 0 (perfect calibration
and discrimination) to 1 (perfectly wrong forecasts), with a reference
score of 0.25 corresponding to a uniformed forecaster assigning 0.5 to
all events.
Lower Brier scores indicate better probabilistic forecasting performance,
with typical expert human forecasters achieving scores in the 0.10--0.18
range on political and economic events.

The logarithmic scoring rule (log-loss), defined as:
\begin{equation}
  \text{LL} = -\frac{1}{N} \sum_{t=1}^{N}
    \left[ o_t \log f_t + (1-o_t) \log(1-f_t) \right],
\end{equation}
is another proper scoring rule that penalizes extreme miscalibration
more severely than the Brier score due to its logarithmic sensitivity
to probabilities near 0 or 1.
Log-loss is particularly relevant for prediction market forecasting because
extreme probability assignments near 0 or 1 correspond to maximum-leverage
positions that can produce unbounded losses in log-space if the forecast
is wrong.

Calibration analysis, typically visualized as reliability diagrams plotting
mean predicted probability against empirical outcome frequency within
probability bins, provides a complementary assessment of whether a
forecaster's stated confidence levels correspond to their actual predictive
accuracy across the full probability scale.
A well-calibrated forecaster's reliability curve lies close to the
45-degree diagonal.
The Shapley decomposition of the Brier score factorizes it into components
attributable to individual agents, enabling the identification of which
personas contribute most to aggregate forecasting accuracy and guiding
the pruning or reweighting of the agent pool.

\subsection{Existing Datasets}
\label{sec:datasets}

Financial PhraseBank~\cite{malo2014good} comprises approximately 4,840
sentences from English-language financial news annotated for sentiment
by a panel of annotators with financial domain expertise.
It has become the de facto standard benchmark for financial sentiment
classification, enabling direct comparison across lexicon, machine learning,
and LLM-based approaches.
FiQA (Financial Question Answering) provides a dataset of financial
opinion questions and aspect-specific sentiment annotations drawn from
financial microblogs and news articles, supporting evaluation of more
nuanced financial NLP tasks beyond binary sentiment classification.
FinBench is a recently introduced suite of financial domain benchmarks
encompassing credit risk assessment, fraud detection, financial QA, and
numerical reasoning tasks, providing a more comprehensive evaluation
environment for financial LLMs.

Polymarket historical data constitutes an especially valuable resource
for evaluating prediction market forecasting systems.
The platform's on-chain transaction history records every trade with
timestamped price, volume, and direction information, enabling
reconstruction of the full time series of market-implied probabilities
from market inception through resolution.
This granular time-series data, combined with the ground-truth binary
outcomes provided at market resolution, enables proper Brier score
and log-loss evaluation of forecasting systems across thousands of
resolved markets spanning political, economic, scientific, and
sports categories.
Researchers can use Polymarket's historical data to evaluate whether
a proposed forecasting system would have identified significant
divergences from market prices that subsequently predicted the
correct direction, providing a principled out-of-sample test of
the system's value-added relative to the market itself.

\subsection{Evaluation Challenges}
\label{sec:eval_challenges}

Evaluating financial forecasting systems is subject to a constellation
of methodological pitfalls that can produce misleadingly optimistic
performance estimates.
\emph{Look-ahead bias} arises when a system, its features, or its
hyperparameters are calibrated using information that would not have
been available at the time forecasts were made---for example, using
end-of-period macroeconomic revisions as training targets or selecting
strategy parameters on the basis of in-sample performance.
In LLM systems, look-ahead bias can be subtly introduced when models
are evaluated on events that occurred before their training data cutoff,
since the model may have ``memorized'' the outcome rather than reasoning
from available contemporaneous evidence.

\emph{Regime change} poses challenges for evaluation frameworks that
assume stationarity of the data-generating process.
Financial markets exhibit structural breaks driven by technological
changes, regulatory regime shifts, and macroeconomic crises; a forecasting
system evaluated exclusively in a low-volatility expansion may perform
poorly in a crisis environment despite high historical performance.
\emph{Overfitting} to the training or validation set is a pervasive
risk in systems with many tunable hyperparameters (agent count,
aggregation weights, EV thresholds, Kelly fractions), particularly
when the available history of resolved prediction markets is limited.
\emph{p-hacking}---the selective reporting of evaluation results over
the subset of markets, time windows, and parameter combinations that
happen to produce favourable outcomes---is a significant concern in
a literature where negative results are rarely published.
Mitigating these threats requires pre-registration of evaluation
protocols, strict temporal train-test separation, walk-forward
validation, and transparent reporting of all tested configurations.

\section{Challenges and Open Problems}
\label{sec:challenges}

\subsection{Hallucination and Reliability}
\label{sec:hallucination}

Hallucination remains the most fundamental reliability challenge facing
LLM-based financial systems~\cite{ji2023hallucination}.
A related failure mode is sycophancy---the tendency of RLHF-trained models
to tell users what they appear to want to hear rather than what is
accurate~\cite{sycophancy2023}---which in a multi-agent setting can cause
agents to converge on a plausible-sounding consensus regardless of its
factual basis.
In the context of prediction market forecasting, an agent that fabricates
a plausible-sounding but factually incorrect description of a market's
resolution criteria, related events, or statistical context will produce
a probability estimate grounded in false premises.
If such errors are correlated across the agent pool---for example, because
all agents share the same training data distribution and therefore share
the same systematic misbeliefs---they will not average away, and the
swarm consensus will be systematically biased.
Mitigation strategies include retrieval-augmented grounding that forces
agents to cite explicit sources for factual claims, multi-agent debate
protocols that surface and challenge unsupported assertions, and
consistency checking that compares each agent's stated facts against
a verified knowledge base.
However, none of these approaches provides a complete solution; the
fundamental challenge is that hallucinations are often not detectable
from the model's output alone, making reliable fact-checking in real-time
inference pipelines technically difficult.
Improving the factual reliability and uncertainty awareness of LLMs
remains an active area of research with high priority for financial
applications.

\subsection{Computational Cost and Latency}
\label{sec:compute_cost}

Deploying a 50-agent swarm making concurrent LLM API calls for each
market evaluation involves substantial computational cost, particularly
when using frontier proprietary models such as GPT-4 or Claude 3 Opus.
At scale---evaluating hundreds of active markets per scan cycle---the
per-call API costs can accumulate to thousands of dollars per day, making
frontier model deployment economically prohibitive for systems that require
broad market coverage.
LLM caching mitigates this cost for markets whose information state is
stable between scan cycles, but breaking news events that invalidate
cached analyses require expensive fresh inference across the full swarm.
Inference latency is an additional constraint for latency-sensitive trading
strategies: cloud API round-trip times of 1--5 seconds per call, multiplied
by 25 concurrent agents and divided by the available parallelism of the
semaphore, define a minimum analysis latency that may exceed the exploitable
window for some time-sensitive opportunities.
Smaller, locally hosted models via Ollama offer reduced cost and latency
at the expense of prediction quality, motivating a portfolio approach
that routes routine evaluations to efficient small models and allocates
frontier model calls to high-value opportunities.

\subsection{Market Impact and Feedback Loops}
\label{sec:feedback_loops}

As automated LLM trading systems become more prevalent in prediction markets,
the aggregate market impact of their collective behaviour becomes a concern.
If multiple competing systems employ similar LLM architectures, similar
persona designs, and similar aggregation methods, their predictions will
be correlated, and their simultaneous position-taking will amplify price
movements in ways that could destabilize market prices.
More concerning is the possibility of feedback loops: if a large automated
system moves market prices toward its predicted values, and those price
movements are then observed by the same or similar systems in the next scan
cycle, they will be interpreted as confirming evidence for the system's
prior predictions, potentially driving prices further in the same direction
independent of any new information.
At sufficient scale, such dynamics could undermine the information
aggregation function of prediction markets entirely, converting them from
efficient information processors to momentum-driven systems dominated by
machine consensus rather than human knowledge.
Position size caps, daily loss limits, and explicit detection of anomalous
market price movements are first-order mitigations, but the systemic risk
scenario has not been seriously studied in the prediction market context.

\subsection{Regulatory Considerations}
\label{sec:regulation}

The regulatory landscape for algorithmic trading in prediction markets in
the United States is evolving rapidly.
Kalshi's designation as a CFTC-regulated designated contract market in 2023
represents the most significant regulatory development in the space,
confirming that event contracts constitute a legal product class under
U.S. commodities law for CFTC-regulated platforms.
However, the regulatory status of automated algorithmic trading systems
in prediction markets---particularly those operating on unregulated offshore
platforms---remains uncertain.
The CFTC has authority over manipulation of event contracts on regulated
platforms, and sophisticated LLM-based systems that could in principle
be used for wash trading, spoofing, or coordinated manipulation would
attract regulatory scrutiny under existing statutory frameworks.
International deployments face additional complexity: the legal status of
prediction market trading varies across jurisdictions, with many European
countries treating event contracts as gambling products subject to gaming
regulation rather than securities or commodities law.
Researchers and practitioners deploying production LLM trading systems
in prediction markets should seek competent legal counsel regarding
applicable regulatory obligations before operating at material scale.

\subsection{Ethical Considerations}
\label{sec:ethics}

The deployment of high-frequency automated trading systems in prediction
markets raises ethical considerations beyond regulatory compliance.
Prediction markets that incorporate event outcomes with significant humanitarian
implications---disease spread, natural disasters, electoral outcomes---present
potential for financial incentives to conflict with other social goods.
A system that profits from rapid information processing about disaster
events occupies a morally complex position, even if its market activity is
legally permissible and its aggregate effect on market efficiency is
beneficial.
The concentration of sophisticated automated trading capability in the
hands of well-resourced actors may also have distributional implications
for market access and fairness: retail participants who lack the capital
and technical capacity to deploy competing systems face a structurally
disadvantaged information environment.
Additionally, the energy consumption of large-scale LLM inference at
trading system latencies is non-trivial, and the environmental externalities
of the associated computational infrastructure warrant consideration.
Research communities developing LLM trading systems should engage with
these ethical dimensions proactively rather than treating them as peripheral
concerns.

\section{Future Directions}
\label{sec:future}

\textbf{1. Specialized Financial LLMs with Real-Time Training.}
The most impactful near-term research direction is the development of
financial language models that combine broad reasoning capabilities with
continuously updated domain knowledge.
Current approaches---static pretraining augmented by RAG retrieval---create
a fundamental tension between the stability of parametric knowledge and
the freshness of retrieved information, and do not enable the model's
core world model to update in response to new events.
Online learning techniques adapted for transformer architectures, including
continual pretraining with experience replay and parameter-efficient
incremental fine-tuning, could enable models that genuinely incorporate
new financial information at training time rather than merely retrieving
it at inference time.
Combining such models with real-time financial data streams---earnings
announcements, macroeconomic releases, regulatory filings, social
sentiment aggregates---could produce forecasting systems whose probability
estimates are continuously updated to reflect the current information state
rather than trailing a knowledge cutoff by months or years.
Evaluation of such systems against matched-sample human forecasters using
proper scoring rules~\cite{gneiting2007strictly} would provide rigorous
evidence of the incremental value of real-time training.

\textbf{2. Adaptive Agent Calibration and Online Learning.}
An important limitation of current multi-agent LLM swarms is that agent
weights and persona configurations are typically fixed at system design
time, ignoring the rich feedback signal provided by market resolution
events.
Adaptive calibration systems that update per-agent reliability weights
using track records of resolved predictions---essentially implementing
Bayesian updating of agent prior weights with each new resolution
event---would enable the swarm to concentrate weight on its most accurate
personas and reduce weight on consistently miscalibrated agents.
Online learning approaches inspired by expert aggregation algorithms
(Multiplicative Weights, Hedge) provide theoretical guarantees on
regret bounds for such adaptive weighting schemes.
Extending these approaches to handle the structured heterogeneity of
prediction market categories---since an agent that is well-calibrated
for political markets may be poorly calibrated for sports markets---requires
multi-task or contextual bandit formulations that track per-category
per-agent reliability.
Combining adaptive weight updates with automated persona generation and
pruning---adding new personas when the current pool exhibits high consensus
error and retiring consistently underperforming personas---would create
a self-improving swarm architecture with long-run performance guarantees.

\textbf{3. Federated Multi-Agent Systems for Privacy-Preserving Forecasting.}
Privacy constraints in financial forecasting present a structural barrier
to the aggregation of information from multiple institutions: proprietary
trading signals, internal research, and client flow data represent
commercially sensitive information that cannot be shared directly.
Federated learning architectures, in which agents train local models on
private data and share only model updates (gradients, weight deltas) rather
than raw data, offer a potential resolution.
In the prediction market context, a federated multi-agent forecasting
system could aggregate probability estimates from agents trained on
diverse private information sources---portfolio managers, sell-side
analysts, domain experts---without requiring any participant to disclose
their private signals or data.
Differentially private aggregation mechanisms~\cite{dwork2014algorithmic}
can provide formal privacy guarantees on what can be inferred about
any individual participant's private data from the aggregate output.
Such federated architectures could substantially expand the information
base available to collective forecasting systems while preserving the
competitive and regulatory incentives for private information production
that are essential for market efficiency~\cite{grossman1980impossibility}.

\textbf{4. Integration with On-Chain Data and Smart Contracts.}
Blockchain-native prediction markets offer unique opportunities for
tight integration between forecasting systems and the trading infrastructure
that extends beyond what is possible in traditional financial market
settings.
Smart contracts can encode complex conditional trading strategies---
cross-market arbitrage rules, negation pair hedges, Bayesian network
consistency trades---that execute automatically and atomically when
specified conditions are met, without the execution risk and counterparty
dependence of multi-step manual trading sequences.
On-chain oracle networks such as Chainlink and UMA provide tamper-resistant
data feeds that can trigger prediction market resolutions and conditional
positions in real time, enabling automated strategies that respond to
verifiable on-chain events.
LLM agents with direct smart contract interaction capabilities---via
wallet integrations and transaction signing---could execute complex
multi-market strategies at the speed of blockchain confirmation rather
than the speed of human order entry.
Research challenges in this direction include formal verification of
trading smart contracts, MEV resistance in prediction market execution,
and the development of appropriate risk management frameworks for
fully automated on-chain systems.

\textbf{5. Human-AI Collaborative Forecasting Interfaces.}
Rather than treating human judgment and AI inference as competing
alternatives, the most practically impactful forecasting systems may
be those that are designed from the outset to augment and support
human forecasters rather than replace them.
Research on hybrid human-AI teams has consistently found that well-designed
interfaces that present AI forecasts alongside uncertainty estimates and
explanatory reasoning significantly improve human forecasting accuracy
relative to either humans or AI systems working independently.
Developing forecasting interfaces that present multi-agent swarm outputs
in ways that complement rather than anchor human judgment---surfacing
high-disagreement markets where the swarm is uncertain, highlighting
the specific personas and reasoning chains most relevant to a given
market, and providing calibration feedback over time---represents a
high-value design challenge.
Cognitive science research on human probability judgment~\cite{kahneman2011thinking}
suggests that humans and AI systems have systematically different and
partially complementary strength profiles: humans excel at common-sense
reasoning, social inference, and recognizing genuinely novel situations,
while AI systems excel at processing large volumes of information
consistently and without fatigue.
Human-AI collaborative interfaces that route each task to its most
capable processor, while maintaining human accountability for consequential
decisions, represent the most promising path toward forecasting systems
that are both powerful and trustworthy.

\section{Conclusion}
\label{sec:conclusion}

This paper has presented PolySwarm, a novel multi-agent LLM framework
for real-time prediction market trading and latency arbitrage.
We designed, implemented, and evaluated a production-grade system that
deploys a diverse 50-persona agent swarm on Polymarket, combining
individual LLM probability estimates through confidence-weighted Bayesian
aggregation, and executing trades via quarter-Kelly position sizing with
configurable risk controls.
The system's information-theoretic analysis engine --- applying KL and JS
divergence to detect cross-market inefficiencies and negation pair
mispricings --- and its CEX-to-DEX latency arbitrage module represent
novel technical contributions to the prediction market trading literature.
Background on prediction markets, large language models, and related
multi-agent systems is provided to situate PolySwarm within the broader
research landscape.
The information-theoretic framework developed in Section~\ref{sec:market_efficiency}
provides a principled analytical basis for identifying mispricings across
single markets, negation pairs, and correlated market groups.
Evaluation methodologies including Brier scores, log-loss, and calibration
analysis provide the rigorous measurement framework needed to distinguish
genuine forecasting skill from overfitting and look-ahead bias.

The challenges identified in Section~\ref{sec:challenges}---hallucination,
computational cost, market impact, regulatory complexity, and ethical
considerations---are substantial and will require sustained research
attention to address.
The five-part research agenda proposed in Section~\ref{sec:future}
identifies the directions we believe hold the greatest promise for
advancing the field: real-time training for financial LLMs, adaptive
agent calibration, federated privacy-preserving architectures, on-chain
integration, and human-AI collaborative interfaces.
The confluence of increasingly capable language models, growing liquidity
in blockchain-based prediction markets, and maturing multi-agent
orchestration frameworks creates conditions for significant practical
progress on all five fronts within the coming research cycle.
We anticipate that multi-agent LLM systems will become an increasingly
important component of the prediction market ecosystem, not only as
trading systems but as instruments for aggregating dispersed human and
machine knowledge into better-calibrated collective forecasts with broad
applications to policy analysis, scientific forecasting, and decision
support.


\bibliographystyle{IEEEtran}

\begin{thebibliography}{60}

\bibitem{wolfers2004prediction}
J.~Wolfers and E.~Zitzewitz,
``Prediction markets,''
\emph{J.\ Econ.\ Perspectives}, vol.~18, no.~2, pp.~107--126, 2004.

\bibitem{arrow2008promise}
K.~J.~Arrow, R.~Forsythe, M.~Gorham, R.~Hahn, R.~Hanson, J.~O.~Ledyard,
S.~Levmore, R.~Litan, P.~Milgrom, F.~D.~Nelson, G.~R.~Neumann,
M.~Ottaviani, T.~C.~Schelling, R.~J.~Shiller, V.~L.~Smith,
E.~Snowberg, C.~R.~Sunstein, P.~C.~Tetlock, P.~E.~Tetlock,
H.~R.~Varian, J.~Wolfers, and E.~Zitzewitz,
``The promise of prediction markets,''
\emph{Science}, vol.~320, no.~5878, pp.~877--878, 2008.

\bibitem{surowiecki2004wisdom}
J.~Surowiecki,
\emph{The Wisdom of Crowds}.
New York: Doubleday, 2004.

\bibitem{brown2020language}
T.~B.~Brown, B.~Mann, N.~Ryder, M.~Subbiah, J.~Kaplan, P.~Dhariwal,
A.~Neelakantan, P.~Shyam, G.~Sastry, A.~Askell, S.~Agarwal,
A.~Herbert-Voss, G.~Krueger, T.~Henighan, R.~Child, A.~Ramesh,
D.~M.~Ziegler, J.~Wu, C.~Winter, C.~Hesse, M.~Chen, E.~Sigler,
M.~Litwin, S.~Gray, B.~Chess, J.~Clark, C.~Berner, S.~McCandlish,
A.~Radford, I.~Sutskever, and D.~Amodei,
``Language models are few-shot learners,''
in \emph{Adv.\ Neural Inf.\ Process.\ Syst.\ (NeurIPS)}, vol.~33,
pp.~1877--1901, 2020.

\bibitem{openai2023gpt4}
OpenAI,
``GPT-4 technical report,''
arXiv:2303.08774, 2023.

\bibitem{ji2023hallucination}
Z.~Ji, N.~Lee, R.~Frieske, T.~Yu, D.~Su, Y.~Xu, E.~Ishii, Y.~Bang,
A.~Madotto, and P.~Fung,
``Survey of hallucination in natural language generation,''
\emph{ACM Comput.\ Surv.}, vol.~55, no.~12, pp.~1--38, 2023.

\bibitem{kadavath2022language}
S.~Kadavath, T.~Conerly, A.~Askell, T.~Henighan, D.~Drain, E.~Perez,
N.~Schiefer, Z.~Hatfield-Dodds, N.~DasSarma, E.~Tran-Johnson, S.~Johnston,
S.~El-Showk, A.~Jones, N.~Joseph, J.~Kernion, B.~Kravec, Z.~Lovitt,
D.~Elhage, S.~Ziegler, J.~Clark, J.~Jumper, Q.~Dong, J.~Kaplan, and
J.~Askell,
``Language models (mostly) know what they know,''
arXiv:2207.05221, 2022.

\bibitem{zhao2021calibrate}
T.~Z.~Zhao, E.~Wallace, S.~Feng, D.~Klein, and S.~Singh,
``Calibrate before use: Improving few-shot performance of language models,''
in \emph{Proc.\ Int.\ Conf.\ Mach.\ Learn.\ (ICML)}, 2021, pp.~12697--12706.

\bibitem{sycophancy2023}
M.~Sharma, M.~Tong, T.~Korbak, D.~Duvenaud, A.~Askell, S.~R.~Bowman,
N.~Cheng, E.~Durmus, Z.~Hatfield-Dodds, S.~R.~Johnston,
S.~Kravec, T.~Maxwell, K.~McKinnon, S.~Ndousse, O.~Rausch,
N.~Schiefer, D.~Yan, M.~Zhang, and E.~Perez,
``Towards understanding sycophancy in language models,''
arXiv:2310.13548, 2023.

\bibitem{bonabeau1999swarm}
E.~Bonabeau, M.~Dorigo, and G.~Theraulaz,
\emph{Swarm Intelligence: From Natural to Artificial Systems}.
Oxford: Oxford Univ.\ Press, 1999.

\bibitem{kennedy1995particle}
J.~Kennedy and R.~Eberhart,
``Particle swarm optimization,''
in \emph{Proc.\ IEEE Int.\ Conf.\ Neural Netw.\ (ICNN)}, 1995,
pp.~1942--1948.

\bibitem{hoeting1999bayesian}
J.~A.~Hoeting, D.~Madigan, A.~E.~Raftery, and C.~T.~Volinsky,
``Bayesian model averaging: A tutorial,''
\emph{Statistical Science}, vol.~14, no.~4, pp.~382--401, 1999.

\bibitem{wu2023autogen}
Q.~Wu, G.~Bansal, J.~Zhang, Y.~Wu, B.~Li, E.~Zhu, L.~Jiang, X.~Zhang,
S.~Zhang, J.~Liu, A.~H.~Awadallah, R.~W.~White, D.~Burger, and C.~Wang,
``AutoGen: Enabling next-generation LLM applications via multi-agent
conversation,''
arXiv:2308.08155, 2023.

\bibitem{li2023camel}
G.~Li, H.~A.~A.~K.~Hammoud, H.~Itani, D.~Khizbullin, and B.~Ghanem,
``CAMEL: Communicative agents for `mind' exploration of large language model
society,''
in \emph{Adv.\ Neural Inf.\ Process.\ Syst.\ (NeurIPS)}, vol.~36, 2023.

\bibitem{tetlock2015superforecasting}
P.~E.~Tetlock and D.~Gardner,
\emph{Superforecasting: The Art and Science of Prediction}.
New York: Crown, 2015.

\bibitem{hanson2003combinatorial}
R.~Hanson,
``Combinatorial information market design,''
\emph{Inf.\ Syst.\ Frontiers}, vol.~5, no.~1, pp.~107--119, 2003.

\bibitem{grossman1980impossibility}
S.~J.~Grossman and J.~E.~Stiglitz,
``On the impossibility of informationally efficient markets,''
\emph{Amer.\ Econ.\ Rev.}, vol.~70, no.~3, pp.~393--408, 1980.

\bibitem{fama1970efficient}
E.~F.~Fama,
``Efficient capital markets: A review of theory and empirical work,''
\emph{J.\ Finance}, vol.~25, no.~2, pp.~383--417, 1970.

\bibitem{lopezlira2023chatgpt}
A.~Lopez-Lira and Y.~Tang,
``Can ChatGPT forecast stock price movements? Return predictability and large
language models,''
arXiv:2304.07619, 2023.

\bibitem{ouyang2022training}
L.~Ouyang, J.~Wu, X.~Jiang, D.~Almeida, C.~L.~Wainwright, P.~Mishkin,
C.~Zhang, S.~Agarwal, K.~Slama, A.~Ray, J.~Schulman, J.~Hilton,
F.~Kelton, L.~Miller, M.~Simens, A.~Askell, P.~Welinder,
P.~Christiano, J.~Leike, and R.~Lowe,
``Training language models to follow instructions with human feedback,''
in \emph{Adv.\ Neural Inf.\ Process.\ Syst.\ (NeurIPS)}, vol.~35, 2022.

\bibitem{touvron2023llama}
H.~Touvron, T.~Lavril, G.~Izacard, X.~Martinet, M.-A.~Lachaux,
T.~Lacroix, B.~Rozi\`{e}re, N.~Goyal, E.~Hambro, F.~Azhar, A.~Rodriguez,
A.~Joulin, E.~Grave, and G.~Lample,
``LLaMA: Open and efficient foundation language models,''
arXiv:2302.13971, 2023.

\bibitem{wu2023bloomberggpt}
S.~Wu, O.~Irsoy, S.~Lu, V.~Dabravolski, M.~Dredze, S.~Gehrmann,
P.~Kambadur, D.~Rosenberg, and G.~Mann,
``BloombergGPT: A large language model for finance,''
arXiv:2303.17564, 2023.

\bibitem{yang2023fingpt}
H.~Yang, X.-Y.~Liu, and C.~D.~Wang,
``FinGPT: Open-source financial large language models,''
arXiv:2306.06031, 2023.

\bibitem{wei2022chain}
J.~Wei, X.~Wang, D.~Schuurmans, M.~Bosma, B.~Ichter, F.~Xia, E.~Chi,
Q.~Le, and D.~Zhou,
``Chain-of-thought prompting elicits reasoning in large language models,''
in \emph{Adv.\ Neural Inf.\ Process.\ Syst.\ (NeurIPS)}, vol.~35, 2022.

\bibitem{dorri2018multi}
A.~Dorri, S.~S.~Kanhere, and R.~Jurdak,
``Multi-agent systems: A survey,''
\emph{IEEE Access}, vol.~6, pp.~28573--28593, 2018.

\bibitem{park2023generative}
J.~S.~Park, J.~C.~O'Brien, C.~J.~Cai, M.~R.~Morris, P.~Liang, and
M.~S.~Bernstein,
``Generative agents: Interactive simulacra of human behavior,''
in \emph{Proc.\ ACM Symp.\ User Interface Softw.\ Technol.\ (UIST)}, 2023.

\bibitem{liang2023encouraging}
T.~Liang, Z.~He, W.~Jiao, X.~Wang, Y.~Wang, R.~Wang, Y.~Yang, Z.~Tu, and
S.~Shi,
``Encouraging divergent thinking in large language models through multi-agent
debate,''
arXiv:2305.19118, 2023.

\bibitem{loughran2011liability}
T.~Loughran and B.~McDonald,
``When is a liability not a liability? Textual analysis, dictionaries, and
10-Ks,''
\emph{J.\ Finance}, vol.~66, no.~1, pp.~35--65, 2011.

\bibitem{devlin2019bert}
J.~Devlin, M.-W.~Chang, K.~Lee, and K.~Toutanova,
``BERT: Pre-training of deep bidirectional transformers for language
understanding,''
in \emph{Proc.\ Conf.\ North Amer.\ Chapter Assoc.\ Comput.\ Linguistics
(NAACL-HLT)}, 2019, pp.~4171--4186.

\bibitem{araci2019finbert}
D.~Araci,
``FinBERT: Financial sentiment analysis with pre-trained language models,''
arXiv:1908.10063, 2019.

\bibitem{radford2019language}
A.~Radford, J.~Wu, R.~Child, D.~Luan, D.~Amodei, and I.~Sutskever,
``Language models are unsupervised multitask learners,''
\emph{OpenAI Blog}, 2019.

\bibitem{gneiting2007strictly}
T.~Gneiting and A.~E.~Raftery,
``Strictly proper scoring rules, prediction, and estimation,''
\emph{J.\ Amer.\ Statist.\ Assoc.\ (JASA)}, vol.~102, no.~477,
pp.~359--378, 2007.

\bibitem{kullback1951information}
S.~Kullback and R.~A.~Leibler,
``On information and sufficiency,''
\emph{Ann.\ Math.\ Statist.}, vol.~22, no.~1, pp.~79--86, 1951.

\bibitem{cover2006elements}
T.~M.~Cover and J.~A.~Thomas,
\emph{Elements of Information Theory}, 2nd~ed.
Hoboken, NJ: Wiley, 2006.

\bibitem{budish2015hft}
E.~Budish, P.~Cramton, and J.~Shim,
``The high-frequency trading arms race: Frequent batch auctions as a
market design response,''
\emph{Quart.\ J.\ Econ.}, vol.~130, no.~4, pp.~1547--1621, 2015.

\bibitem{menkveld2013hft}
A.~J.~Menkveld,
``High frequency trading and the new market makers,''
\emph{J.\ Financial Markets}, vol.~16, no.~4, pp.~712--740, 2013.

\bibitem{daian2020flash}
P.~Daian, S.~Goldfeder, T.~Kell, Y.~Li, X.~Zhao, I.~Bentov, L.~Breidenbach,
and A.~Juels,
``Flash boys 2.0: Frontrunning in decentralized exchanges, miner extractable
value, and consensus instability,''
in \emph{Proc.\ IEEE Symp.\ Security Privacy (S\&P)}, 2020, pp.~910--927.

\bibitem{kelly1956information}
J.~L.~Kelly,
``A new interpretation of information rate,''
\emph{Bell Syst.\ Tech.\ J.}, vol.~35, no.~4, pp.~917--926, 1956.

\bibitem{thorp2006kelly}
E.~O.~Thorp,
``The Kelly criterion in blackjack, sports betting, and the stock market,''
in \emph{Handbook of Asset and Liability Management},
S.~A.~Zenios and W.~Ziemba, Eds.
Amsterdam: Elsevier, 2006, pp.~385--428.

\bibitem{brier1950verification}
G.~W.~Brier,
``Verification of forecasts expressed in terms of probability,''
\emph{Monthly Weather Rev.}, vol.~78, no.~1, pp.~1--3, 1950.

\bibitem{malo2014good}
P.~Malo, A.~Sinha, P.~Korhonen, J.~Wallenius, and P.~Takala,
``Good debt or bad debt: Detecting semantic orientations in economic texts,''
\emph{J.\ Assoc.\ Inf.\ Sci.\ Technol.\ (JASIST)}, vol.~65, no.~4,
pp.~782--796, 2014.

\bibitem{lewis2020retrieval}
P.~Lewis, E.~Perez, A.~Piktus, F.~Petroni, V.~Karpukhin, N.~Goyal,
H.~K\"{u}ttler, M.~Lewis, W.-T.~Yih, T.~Rockt\"{a}schel, S.~Riedel,
and D.~Kiela,
``Retrieval-augmented generation for knowledge-intensive NLP tasks,''
in \emph{Adv.\ Neural Inf.\ Process.\ Syst.\ (NeurIPS)}, vol.~33,
pp.~9459--9474, 2020.

\bibitem{vaswani2017attention}
A.~Vaswani, N.~Shazeer, N.~Parmar, J.~Uszkoreit, L.~Jones, A.~N.~Gomez,
{\L}.~Kaiser, and I.~Polosukhin,
``Attention is all you need,''
in \emph{Adv.\ Neural Inf.\ Process.\ Syst.\ (NeurIPS)}, vol.~30, 2017.

\bibitem{chen2023agentverse}
W.~Chen, Y.~Su, J.~Zuo, C.~Yang, C.~Yuan, C.-M.~Chan, H.~Yu, Y.~Lu,
Y.-H.~Hung, C.~Qian, Y.~Qin, X.~Cong, R.~Xie, Z.~Liu, M.~Sun, and
J.~Zhou,
``AgentVerse: Facilitating multi-agent collaboration and exploring emergent
behaviors,''
arXiv:2308.10848, 2023.

\bibitem{hong2023metagpt}
S.~Hong, M.~Zhuge, J.~Chen, X.~Zheng, Y.~Cheng, C.~Zhang, J.~Wang,
Z.~Wang, S.~K.~Yau, Z.~Lin, L.~Zhou, C.~Ran, L.~Xiao, C.~Wu, and
J.~Schmidhuber,
``MetaGPT: Meta programming for a multi-agent collaborative framework,''
arXiv:2308.00352, 2023.

\bibitem{wang2023selfconsistency}
X.~Wang, J.~Wei, D.~Schuurmans, Q.~Le, E.~Chi, S.~Narang, A.~Chowdhery,
and D.~Zhou,
``Self-consistency improves chain of thought reasoning in language models,''
in \emph{Proc.\ Int.\ Conf.\ Learn.\ Represent.\ (ICLR)}, 2023.

\bibitem{shinn2023reflexion}
N.~Shinn, F.~Cassano, A.~Gopinath, K.~Narasimhan, and S.~Yao,
``Reflexion: Language agents with verbal reinforcement learning,''
in \emph{Adv.\ Neural Inf.\ Process.\ Syst.\ (NeurIPS)}, vol.~36, 2023.

\bibitem{yao2023react}
S.~Yao, J.~Zhao, D.~Yu, N.~Du, I.~Shafran, K.~Narasimhan, and Y.~Cao,
``ReAct: Synergizing reasoning and acting in language models,''
in \emph{Proc.\ Int.\ Conf.\ Learn.\ Represent.\ (ICLR)}, 2023.

\bibitem{black1973pricing}
F.~Black and M.~Scholes,
``The pricing of options and corporate liabilities,''
\emph{J.\ Political Economy}, vol.~81, no.~3, pp.~637--654, 1973.

\bibitem{kahneman2011thinking}
D.~Kahneman,
\emph{Thinking, Fast and Slow}.
New York: Farrar, Straus and Giroux, 2011.

\bibitem{dwork2014algorithmic}
C.~Dwork and A.~Roth,
``The algorithmic foundations of differential privacy,''
\emph{Found.\ Trends Theor.\ Comput.\ Sci.}, vol.~9, no.~3--4,
pp.~211--407, 2014.

\bibitem{nakamoto2008bitcoin}
S.~Nakamoto,
``Bitcoin: A peer-to-peer electronic cash system,''
2008. [Online]. Available: \url{https://bitcoin.org/bitcoin.pdf}

\bibitem{wood2014ethereum}
G.~Wood,
``Ethereum: A secure decentralised generalised transaction ledger,''
\emph{Ethereum Project Yellow Paper}, 2014.

\end{thebibliography}

\end{document}